\documentclass[a4paper]{article}
\usepackage{INTERSPEECH2022}
\usepackage{xcolor}
\usepackage{multirow}
\usepackage[colorlinks,linkcolor=blue]{hyperref}
\usepackage{enumitem}
\usepackage{verbatim}

\title{ASR Error Correction with Constrained Decoding on Operation Prediction}
\name{Jingyuan Yang\textsuperscript{$\dagger$}, Rongjun Li\textsuperscript{$\dagger$}, Wei Peng*
\thanks{\textdagger Indicating equal contribution}
\thanks{*Corresponding author}}
\address{
  Artificial Intelligence Application Research Center, Huawei Technologies}
\email{yangjingyuan2@huawei.com, lirongjun3@huawei.com, peng.wei1@huawei.com}

\begin{document}

\maketitle
\begin{abstract}
Error correction techniques remain effective to refine outputs from automatic speech recognition (ASR) models. Existing end-to-end error correction methods based on an encoder-decoder architecture process all tokens in the decoding phase, creating undesirable latency. In this paper, we propose an ASR error correction method utilizing the predictions of correction operations. More specifically, we construct a predictor between the encoder and the decoder to learn if a token should be kept (``K"), deleted (``D"), or changed (``C") to restrict decoding to only part of the input sequence embeddings (the ``C" tokens) for fast inference. Experiments on three public datasets demonstrate the effectiveness of the proposed approach in reducing the latency of the decoding process in ASR correction. It enhances the inference speed by at least three times (3.4 and 5.7 times) while maintaining the same level of accuracy (with WER reductions of 0.53\% and 1.69\% respectively) for our two proposed models compared to a solid encoder-decoder baseline. In the meantime, we produce and release a benchmark dataset contributing to the ASR error correction community to foster research along this line. 

\end{abstract}

\section{Introduction}
Automatic speech recognition (ASR) models play critical roles in pipeline-based contemporary deep learning systems (i.e., Voice Assistant and Speech Translation) by transforming sequences of audio signals to sequences of word tokens for down-streaming tasks \cite{huang2020learning,weng2020joint}. Despite its comprehensive application to industrial scenarios, ASR is ill-performed in real-world cases, whereas out-of-domain speech and noisy data prevail. For example, Table \ref{tab:examples0} shows several common ASR errors, including an entity error with a reference data ``to orlando" wrongly recognized as ``tewel ando", leading to problems in down-streaming tasks. 

Error correction (EC) techniques remain effective means to refine outputs from ASR models. These EC approaches belong to a cascade framework and an end-to-end sequence to sequence framework. The cascade framework consists of a sequence labeling method to identify the sequence of interests, followed by a rule-based value error recovery module \cite{liu2020robust}. A cascading EC framework requires costly manual transcriptions to all positive and negative tokens; therefore, scaling up a cascading EC approach is challenging. An end-to-end framework treats EC of ASR as a machine translation method between problematic sequences of tokens and reference sequences \cite{guo2019spelling,wang2020asr,bekal2021remember,zhao2021bart}. Along this line, Transformer-based ASR error correction methods learn a mapping between the outputs of ASR and the related ground-truth transcriptions \cite{zhang2019investigation}. Pretrained language models, i.e., BERT \cite{kenton2019bert} and BART \cite{lewis2019bart}, are used as encoders to represent input sequences to boost ASR error correction \cite{li2021boost}, in which a Bart-initialized model is regarded to significantly outperform that from BERT \cite{zhao2021bart}. The end-to-end ASR error correction framework becomes the mainstream as it achieves the state-of-the-art (SOTA) WER and requires fewer efforts in preparing training data than the cascading framework. Existing transformer-based ASR EC models process all input embeddings equally in the decoding phase, creating undesirable latency. In this paper, we propose an ASR error correction method utilizing the predictions of correction operations to constrain the decoding and reduce the inference latency. More specifically, we design a predictor between the encoder and the decoder to classify if a token should be kept (``K"), deleted (``D") or changed (``C") to restrict decoding to only part of the input sequence embeddings (the ``C" tokens) for fast inference. On the other hand, there is a lack of publicly available benchmark datasets for ASR error correction as researchers in this area majorly report the experimental results on the in-house data. Such data deficiency in the public domain has inhibited the advancement of the ASR error correction techniques. The contributions of our work are two folds:
\begin{itemize}[leftmargin=*]
\item To our best knowledge, we are the first to construct an operation predictor to perform constrained decoding to enhance the inference speed of an ASR error correction approach. Experiments on three datasets demonstrate the effectiveness of the proposed method in enhancing the inference speed by at least three times while maintaining the same level of accuracy (WER) compared to a solid encoder-decoder baseline.   
\item To address the lack of public benchmark datasets for ASR error correction, we apply internal ASR and Text-to-Speech (TTS) engines to a single-turn dialogue dataset TOP \cite{gupta2018semantic}. In this way, we produce an ASR error correction dataset contributing to the community to foster research along this line.         
\end{itemize}

The code and datasets\footnote{\url{https://github.com/yangjingyuan/ConstDecoder}} are made publicly available.

\begin{table}[t]
  \caption{Typical ASR errors include: 1) \textcolor{red}{the grammatical error due to unclear voice signal (in red fonts)}; 2) \textcolor{blue}{the similar sound error created by phonetically confusing words (in blue)}; 3) \textcolor{olive}{the lack of domain-specific terms causing the entity error (in olive fonts) }; 4) \textcolor{cyan}{ the insertion error in cyan}, \textcolor{orange}{and the delete error in orange}  }
  \label{tab:examples0}
  \centering
  \begin{tabular}{l|l}
  \toprule
  Type & Example \\
  \midrule
  \multirow{2}{*}{Reference} 
  & \textcolor{red}{cheapest} \textcolor{blue}{airfare} from tacoma \textcolor{olive}{to orlando}.\\
  & will it get \textcolor{cyan}{*} hotter in \textcolor{orange}{hext}.  \\
  \hline
  \multirow{2}{*}{ASR}    
  & \textcolor{red}{cheaper} \textcolor{blue}{stair} from tacoma \textcolor{olive}{tewel ando}.  \\
  & will it get \textcolor{cyan}{a} hotter in \textcolor{orange}{*}.\\
  \bottomrule
  \end{tabular}
\end{table}

\begin{figure*}[ht]
  \centering
  \includegraphics[width=0.7\linewidth]{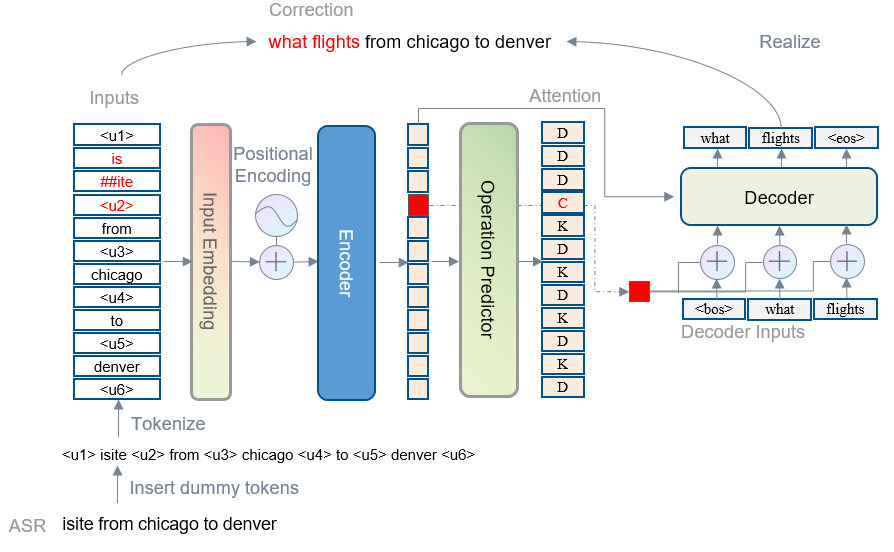}
  \caption{The overall architecture of the proposed method. The decoder takes the encoded hidden representation over the ``C'' positions, the attended context vector, and the prediction from the last time step as inputs to produce the corrected tokens. In particular,  when multiple ``C'' tags are present, they are decoded in parallel.}
  \label{fig:architecture}
\end{figure*}

\section{Related Work}
ASR correction has been jointly trained with natural language understanding in a multi-task setting \cite{weng2020joint}. To address the issue associated with semantic-only post-correction, phonetic features are included in the correction model to tackle homophonic errors commonly identified in Chinese ASR \cite{chen2021integrated}. Along this direction, \cite{wang2020asr} propose a variant transformer model encoding both the word and phoneme sequence of an entity to fix name entity errors. A similar research \cite{serai2022hallucination}  utilizes a dual encoder in an end-to-end framework to handle an input word sequence and corresponding phoneme sequence. More works \cite{li2021boost,zhao2021bart} have emerged to boost the accuracy of error correction. It was not until recently that researchers commenced exploring means to reduce decoding latency associated with sequence-to-sequence ASR correction models.
LaserTagger \cite{malmi2019encode} converts the sequence generation task to a text editing one to enhance inference speed. However, LaserTagger can not handle the out-of-vocabulary problem as it relies on a predefined vocabulary. Although there is an attempt leveraging non-autoregressive models in ASR correction, the study \cite{leng2021fastcorrect1} is built on a large internal Mandarin speech corpus. It is argued that an error correction method using non-autoregressive models suffer from the issue of over-correction, resulting in mis-replacement of normal words. A direct comparison is not applicable before a full-scale release of code and data. \cite{huang2021sarg} propose a semi-autoregressive model coupled with an LSTM for generation, achieving SOTA accuracy and inference speed in the utterance restoration. Inspired by \cite{huang2021sarg}, we design an end-to-end ASR error correction method using an operation predictor to restrict decoding to only desirable parts of the input sequence embeddings to reduce inference latency.         

\section{Approach}
The task of ASR error correction can be treated as a functional mapping $f: X \to T$ between $X$ and $T$, where $X$ denotes text outputs from ASR containing errors, $T$ is the ground-truth reference text. Unlike other sequence generation tasks, which may have radically different inputs and outputs, it is observed that most contents of $X$ and $T$ are identical. The intuition here is to only apply a mapping ($f$) between the desirable part of $X$ and the counterparts in $T$. We design an end-to-end architecture consisting of an encoder, a decoder, and an operation predictor to constrain the decoding process to the desired part of the input sequence, shown in Figure \ref{fig:architecture}. The encoder produces sequence embeddings from the word embeddings of the ASR output texts and related positional embeddings. The operation predictor is a fully-connected layer trained to identify the above-mentioned ``K", ``D" and ``C" input sequence embeddings. The model can reduce inference latency significantly by restricting the decoding process to the ``C" tokens.             
\subsection{Preprocessing phase}
The preprocessing step is similar to that mentioned in \cite{huang2021sarg}:(1) Insert dummy tokens between very adjacent token in $X$; (2) Apply WordPiece tokenizer \cite{wu2016google} to $X$ and $T$ to produce corresponding token sequences $X=\{\omega_{1}^{x},\omega_{2}^{x},...,\omega_{n}^{x}\}$, $T=\{\omega_{1}^{t},\omega_{2}^{t},...,\omega_{m}^{t}\}$;
(3) Construct training labels for the operation predictor and the decoder. We calculate the longest common subsequence (LCS) between $X$ and $T$, and apply LCS to align $X$ and $T$ recurrently.
The aligned tokens are labeled as ``KEEP" (``K") with the rest tagged as ``DELETE" (``D") or ``CHANGE" (``C") with the positions of the chunks of ``C" tokens recorded. 
The operation training labels are defined as $D=\{d_{1},d_{2},...,d_{n}\}$. The decoding sequence for ``C" commencing at $k^{th}$ position can be denoted as $Y_{k}=\{\omega_{1,k}^{y},\omega_{2,k}^{y},...,\omega_{o,k}^{y}\}$.       

\subsection{Model architecture}
We use BERT as an encoder to obtain the hidden representations of the model inputs $X$:
\begin{align}
E^{x} &= \mathrm{BERT}_{\mathrm{Encoder}}(\mathrm{WE}(X) + \mathrm{PE}(X))
\label{eq.1}
\end{align}
where $\mathrm{WE}$ and $\mathrm{PE}$ are related word and position embedding functions respectively. $E^{x}= \{e_{1}^{x},e_{2}^{x},...,e_{n}^{x}\}$ is the encoded representation of $X$, and $e_{i}^{x}$ is the $i^{th}$ encoded token representation of $X$. Next, one fully-connected layer with softmax is designed to be the operation predictor:
\begin{align}
p(\hat{d_{i}}|\omega_{i}^{x}) &= \mathrm{softmax}(W_{d}e_{i}^{x} + b_{d}).
\end{align}
Here, $\hat{d_{i}}$ is the predicted operation. $W_{d}$ and $b_{d}$ are the learnable parameters.
At last, the operation loss is defined as negative log-likehood:
\begin{align}
\mathrm{loss}_{oper} &= -\sum_{i} \mathrm{log}(p(\hat{d_{i}}|\omega_{i}^{x})).
\end{align}

Two kinds of constrained decoders are implemented, one is LSTM-based and the other is transformer-based.

\begin{itemize}
\item[$\bullet$] \textbf{LSTM-based decoder}
\end{itemize}
\begin{align}
s_{t+1}^{k} &= \mathrm{LSTM}( \mathrm{WE}(\omega_{t,k}^{y}), s_{t}^{k}).
\end{align}

where $\omega_{t,k}^{y}$ and $s_{t}^{k}$ are the decoder input and the LSTM hidden state at time step $t$ in terms of the $k^{th}$ change operation. The initial values of $\omega_{0,k}^{y}$ and $s_{0}^{k}$ are $<$bos$>$ and $e_{k}^{x}$ respectively. WE is the same word embedding function as in equation \ref{eq.1}.
Moreover, corresponding context vector is calculated based on the additive attention \cite{bahdanau2015neural} mechanism:
\begin{align}
&c_{t+1}^{k} = \mathrm{Attn}(Q, K, V),\\
&Q =s_{t+1}^{k}, K=E^{x}, V=E^{x}.
\end{align}

where the LSTM hidden state $s_{t+1}^{k}$ works as query (``Q"), and the encoded model input $E^x$ works as key (``K") and value (``V") in corresponding attention calculation.
Finally, the context vector $c_{t+1}^{k}$ and the LSTM hidden state $s_{t+1}^{k}$ are fused to predict the target text.
\begin{align}
p(\hat{\omega}_{t+1,k}^{y}) = \mathrm{softmax}(W_{y} [c_{t+1}^{k} \oplus s_{t+1}^{k}] + b_{y})
\end{align}

where $\oplus $ indicates a concatenate function, $W_y$ and $b_y$ are the learnable parameters.

\begin{itemize}
\item[$\bullet$] \textbf{Transformer decoder}
\end{itemize}

For the $k^{th}$ change position, the transformed-based decoder inputs at timestep $t$ are calculated as follows:  
\begin{align}
E_{t,k}^{y} &= W_{p} [ (\mathrm{WE}(\omega_{t,k}^{y}) + \mathrm{PE}(\omega_{t,k}^{y})) \oplus e_{k}^{x}].
\end{align}

where WE and PE are the same word and position embedding functions as in equation \ref{eq.1}. $e_{k}^{x}$ is the encoded token representation mentioned before. $W_{p}$ is the trainable parameters.
Then, a standard transformer decoder \cite{vaswani2017attention} is employed to fuse all relevant information:
\begin{align}
h_{t+1}^{k} &= \mathrm{TransformerDecoder}(Q, K, V),\\
Q &= E_{t,k}^{y}, K = E^{x},V = E^{x}.
\end{align}
where the decoder input $E^{y}_{t,k}$ works as query (``Q"), and encoded model input $E^{x}$ works as key (``K") and value (``V") in transformer decoder. $h_{t+1}^{k}$ is the fused vector.
Finally, the target output is predicted as follows:
\begin{align}
p(\hat{\omega}_{t+1,k}^{y}) = \mathrm{softmax}(W_{y} h_{t+1}^{k}+ b_{y})
\end{align}
where $W_y$ and $b_y$ are trained parameters.
Both LSTM-based and transformer-based generation loss are defined as:
\begin{align}
\mathrm{loss}_{dec} &= -\sum_{k}\sum_{t} \mathrm{log}(p(\hat{\omega}_{t,k}^{y})).
\end{align}

The total loss is calculated as follows:
\begin{align}
\mathrm{loss} &= \alpha \mathrm{loss}_{oper} + \mathrm{loss}_{dec}.
\end{align}
where $\alpha$ is the hyperparameter.

\section{Experiments}
In order to evaluate the effectiveness of our proposed model, three datasets leveraging different ASR engines are utilized.  
\begin{table}[htb]
  \caption{Statistics of the utilized datasets}
  \label{tab:data_info}
  \centering
  \begin{tabular}{l|rrr}
    \toprule
    \textbf{Name} & \textbf{Train} & \textbf{Valid} & \textbf{Test}    \\
    \midrule
    ATIS	&   3867	&   967 &   800	\\
    SNIPS   &   13084   &   700 &   700 \\
    TOP &   31279	&   4462    &   9042 \\
    \bottomrule
  \end{tabular}
\end{table}

\begin{figure}[htb]
  \centering
  \includegraphics[width=0.9\linewidth]{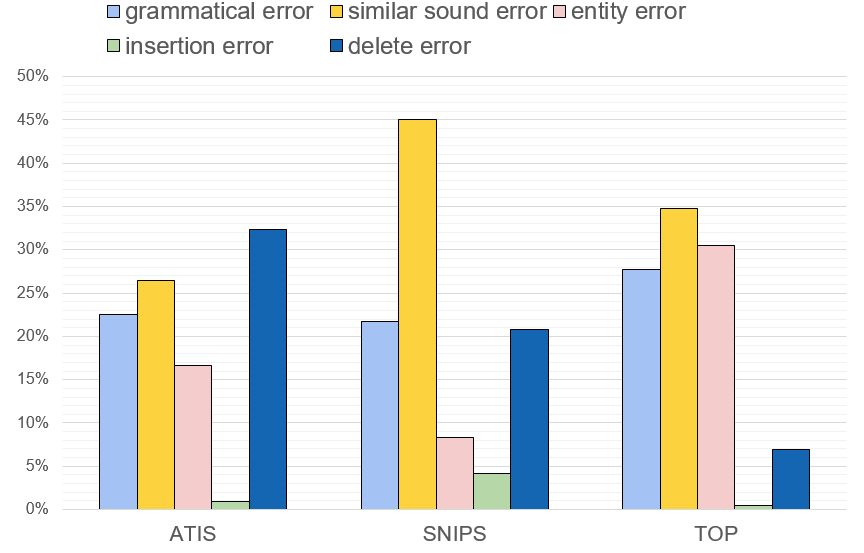}
  \caption{A comparison of ASR error type distributions for three benchmark datasets.}
  \label{fig:asr_errors}
\end{figure}

\begin{itemize}[leftmargin=*]
\item ATIS \cite{hemphill1990atis} is a public benchmark dataset containing user voice records of flight information. \cite{sundararaman2021phoneme} used a TTS engine (Amazon Polly\footnote{\url{https://aws.amazon.com/polly/}}) to convert the text to voice, adding ambient noise\footnote{\url{www.pacdv.com/sounds/ambience_sounds.html}} to simulate real-life data.  Finally, a LAS \cite{ChanJLV16} ASR engine was used to convert voice data into transcribed texts. 

\item SNIPS \cite{coucke2018snips} is also a benchmark dataset for speaking language understanding study. \cite{huang2020learning} used a commercial TTS to synthesize audio data from texts. Kaldi\footnote{\url{https://github.com/kaldi-asr/kaldi}} is used to transcribe audio to texts. 

\item TOP \cite{gupta2018semantic} is the dataset proposed by Facebook focusing on the topics of navigation and events. We use our own commercial TTS and ASR engine to produce the dataset. 
\end{itemize}

\begin{table}[htb]
  \caption{The experimental results of our method against other SOTA methods in WER on three benchmark datasets. ConstDecoder$_{lstm}$ and ConstDecoder$_{trans}$ refer to our proposed constrained decoder using LSTM and transformer respectively. }
  \label{tab:main_result}
  \centering
  \begin{tabular}{l|rrr|r}
    \toprule
    \textbf{Method} & \textbf{ATIS} & \textbf{SNIPS} & \textbf{TOP} & \textbf{Average}\\
    \midrule
    Original & 30.65 & 45.73 & 13.45 & 29.94 \\
    SC\_BART \cite{zhao2021bart} & 21.47 & 30.35 & 7.30 & 19.70\\
    distilBART \cite{shleifer2020pre} & 26.51 & 33.28 & 8.25 & 22.68 \\
    \hline
    ConstDecoder$_{lstm}$ & 22.44 & 31.75 & 10.02 & 21.40\\
    vs. Original  & -8.21 & -13.98 & -3.43 & -8.54 \\
    vs. SC\_BART & +0.97 & +1.40 & +2.72 & +1.69 \\
    vs. distilBART & -4.07 & -1.53 & +1.77 & -1.28 \\

    \hline
    ConstDecoder$_{trans}$ & 21.74 & 30.98 & 7.99 & 20.23\\
    vs. Original  & -8.91 & -14.75 & -5.46 & -9.70 \\
    vs. SC\_BART & +0.27 & +0.63 & +0.69 & +0.53 \\
    vs. distilBART & -4.77 & -2.30 & -0.26 & -2.44\\
    \bottomrule
  \end{tabular}
\end{table}

\begin{table}[htb]
  \caption{The inference time cost in milliseconds (ms) against a SOTA transformer-based error correction model (SC\_BART) and a distilled version of it (distilBART) on three benchmark datasets. }
  \label{tab:speed}
  \centering
  \begin{tabular}{l|rrr|r}
    \toprule
    \textbf{Method} & \textbf{ATIS} & \textbf{SNIPS} & \textbf{TOP} & \textbf{Average} \\
    \midrule
    SC\_BART & 90.30 & 75.30 & 75.08 & 80.22\\
    distilBART & 45.55 & 41.55 & 40.08 & 42.39 \\
    \hline
    ConstDecoder$_{lstm}$ & 14.31 & 15.06 & 12.53 & 13.96   \\
    vs. SC\_BART & 6.3$\times$ & 5.0$\times$ & 6.0$\times$ & 5.7$\times$\\
    vs. distilBART & 3.2$\times$ & 2.8$\times$ & 3.2$\times$ & 3.0$\times$ \\
    \hline
    ConstDecoder$_{trans}$ & 25.61 & 26.66 & 18.43 & 23.56\\
    vs. SC\_BART & 3.5$\times$ & 2.8$\times$ & 4.1$\times$ & 3.4$\times$ \\
    vs. distilBART & 1.8$\times$ & 1.6$\times$ & 2.2$\times$ & 1.8$\times$ \\
    \bottomrule
  \end{tabular}
\end{table}

Figure \ref{fig:asr_errors} shows a distribution of errors in our sampled datasets. It is noted that they are significantly different, enabling a comprehensive evaluation of model effectiveness. We compare the propose method with the following three baselines:
\begin{itemize}[leftmargin=*]
\item Original: refers to the WER of the original datasets when no error correction applies;
\item SC\_BART \cite{zhao2021bart}: a pretrained encoder-decoder transformer language model, which has reached SOTA results on ASR error correction tasks;
\item distillBART \cite{shleifer2020pre}: a distilled version of BART-large, which has a 12-layer encoder and a single-layer decoder. The application of distillation followed by finetuning on the three datasets to enhance the inference speed.
\end{itemize}

\subsection{Model training details}
The encoder of the ConstDecoder$_{lstm}$ is initialized using a pretained BERT (bert-base-uncased) with 12-layer architecture and a hidden size of 768. The decoder is a single-layer LSTM with a hidden size of 768. The optimizer is AdamW \cite{LoshchilovH19} with a batch size of 8, trained with 20 epochs. The encoder of ConstDecoder$_{trans}$ adopts the same setting with the that of ConstDecoder$_{lstm}$. The decoder is a single-layer transformer decoder with a hidden size of 768. We use Adam \cite{KingmaB14} as the optimizer with a batch size of 32 (the same as baseline), trained with 20 epochs. The maximum length of the decoder output is set to 10. The coefficient ($\alpha$) of the loss function is initialized to 3. The initial learning rate is 5e-5. The model is trained and evaluated using NVIDIA Tesla V100.   

\subsection{Results analysis}

Tables \ref{tab:main_result}-\ref{tab:speed} show the experimental results in terms of WER and inference speed of the proposed method against a SOTA transformer baseline (SC\_BART) on three benchmark datasets. Both SC\_BART and our method can reduce WER for the original datasets significantly. SC\_BART performs slightly better than our method (0.53\%) in WER, possibly ascribing to: 1) It utilizes a more robust pretrained language model and; 2) It engages a fully autoregressive model. However, our method leads in inference speed, achieving at least three times faster decoding time on average for the three involved benchmark datasets. Applying model distillation techniques to BART-large followed by finetuning can enhance its inference speed moderately to 42.39 milliseconds but produce a worse WER (2.44\% lower than that of ConstDecoder$_{trans}$).

It can be observed that a transformer is a preferable option to LSTM for implementing the decoder in our method when WER is a major concern. Furthermore, our method achieves consistent results in all three benchmark datasets produced by different ASR techniques, demonstrating a reasonable level of robustness.    

\begin{table}[htb]
  \caption{The effectiveness of the corrected ASR transcribed texts for different error types. For example, out of 61 grammatical errors in the original samples, 45 are corrected, resulting in a correction ratio (Cor. Ratio) of 73.77\%.}
  \label{tab:corrections}
  \centering
  \begin{tabular}{l|l}
  \toprule
    \textbf{Error Type} & \textbf{Example}          \\
    (Cor. Ratio) & (ASR vs. Correction)          \\
  \midrule
  grammatical & ASR: which \textcolor{red}{flight sleep} chicago on april \\
  (73.77\%) & Cor.: which flights \textcolor{red}{leave} chicago on april \\
  \hline
  similar sound & ASR: sport events in \textcolor{red}{roll e} this weekend\\
  (32.08\%) & Cor.: sporting events in \textcolor{red}{raleigh} this weekend\\
  \hline
  insertion & ASR: arrive in \textcolor{red}{indian} indianapolis \\
  (16.67\%) & Cor.: arrive in indianapolis \\
  \hline
  entity & ASR: from \textcolor{red}{mil} to orlando \\
  (10.20\%) & Cor.: from \textcolor{red}{milwaukee} to orlando \\ 
\hline
  delete & ASR: most of the movie \\
  (3.17\%) & Cor.: what are the movie \textcolor{red}{schedules}\\
  \bottomrule
  \end{tabular}
\end{table}

Table \ref{tab:corrections} illustrates the result of a sampled analysis for the correction ratio presented in three datasets by randomly sampling one hundred cases from each dataset. It can be observed that the proposed method can address grammatical errors (73.77\%) and issues associated with similar sounds (32.08\%) effectively. A big gap is identified in addressing the entity, insertion, and deletion errors. A further look into the negative cases discloses that entity errors may be caused by a lack of knowledge handling the out-of-vocabulary (OOV) problem. It is worth noting that integrating a memory module to capture rare entities incrementally during error correction may be the solution, and we leave this as future work. 

\section{Conclusions}
In this paper, we illustrate an end-to-end ASR error correction method leveraging constrained decoding on operation prediction. Experiments on three benchmark datasets demonstrate the effectiveness of the proposed models in boosting inference speed by at least three times (3.4 and 5.7 times) while maintaining the same level of accuracy (with WER reductions of 0.53\% and 1.69\% respectively) compared to a SOTA transformer-based baseline. In the meantime, we apply internal ASR and TTS engines to a single-turn dialogue dataset, turning it into a publicly available benchmark dataset for ASR error correction. Future work will investigate the roles of incremental learning mechanisms in practice-oriented ASR error correction approaches to handle issues associated with the OOV problem mentioned above. 

\bibliographystyle{IEEEtran}
\bibliography{submit}
\end{document}